\documentclass[letterpaper, 10 pt, conference]{ieeeconf}  

\IEEEoverridecommandlockouts                              
\usepackage[utf8]{inputenc}
\usepackage[T1]{fontenc}
\usepackage{verbatim}
\usepackage{bm}
\usepackage{amsmath}
\usepackage{graphicx}
\usepackage{caption2}
\usepackage{subfigure}
\usepackage{algorithm}
\usepackage{algorithmic}
\usepackage{multirow}
\usepackage{hyperref}
\usepackage{float}
\usepackage{flushend}
\usepackage{amsmath}

\title{\LARGE \bf
KinoJGM: A framework for efficient and accurate quadrotor trajectory generation and tracking in dynamic environments
}

\author{Yanran Wang, James O'Keeffe, Qiuchen Qian and David Boyle
\thanks{This work was supported by the UK Government under NSF-UKRI Grant NE/T011467/1. The authors are with the Dyson School of Design Engineering, Imperial College London, United Kingdom. {\tt\small yanran.wang20, j.okeeffe, qiuchen.qian19, david.boyle at imperial.ac.uk}}%
}

\begin{document}

\maketitle
\thispagestyle{empty}
\pagestyle{empty}

\begin{abstract}

Unmapped areas and aerodynamic disturbances render autonomous navigation with quadrotors extremely challenging. To fly safely and efficiently, trajectory planners and trackers must be able to navigate unknown environments with unpredictable aerodynamic effects in real-time. 
When encountering aerodynamic effects such as strong winds, most current approaches to quadrotor trajectory planning and tracking will not attempt to deviate from a determined plan, even if it is risky, in the hope that any aerodynamic disturbances can be resisted by a robust controller.
This paper presents a novel systematic trajectory planning and tracking framework for autonomous quadrotors. We propose a Kinodynamic Jump Space Search (Kino-JSS) to generate a safe and efficient route in unknown environments with aerodynamic disturbances. A real-time Gaussian Process is employed to model the errors caused by aerodynamic disturbances, which we then integrate with a Model Predictive Controller to achieve efficient and accurate trajectory optimization and tracking.
We demonstrate our system to improve the efficiency of trajectory generation in unknown environments by up to 75\% in the cases tested, compared with recent state-of-the-art. We also show that our system improves the accuracy of tracking in selected environments with unpredictable aerodynamic effects. Our implementation is available in an open source package\footnote{\url{https://github.com/Alex-yanranwang/Imperial-KinoJGM}}.

\end{abstract}

\section{INTRODUCTION}
It is crucial that autonomous Unmanned Aerial Vehicles (UAVs), such as quadrotors, can maintain their autonomy in complex and dynamic environments. Achieving safe, efficient and reliable autonomous navigation for quadrotors in GPS-denied environments, unmapped environments and/or those with large aerodynamic disturbances - e.g., strong wind or turbulence - is an ongoing challenge with considerable significance to industrial applications, such as commercial deliveries, search-and-rescue and smart agriculture. 

Most existing methods on quadrotor trajectory generation and tracking have so far been demonstrated in free known space (free space here refers to unoccupied space) \cite{tordesill2021faster,ryll2019efficient,mellinger2011minimum}. For an unknown environment and environments in which aerodynamic disturbances can jeopardize the stability of a quadrotor there are two main problems that need to be solved: How can an optimized trajectory be generated efficiently in real-time whilst guaranteeing safety and feasibility? And; How can a feasible trajectory be generated and tracked using limited onboard computational resources?

Generating optimized trajectories for quadrotors is typically solved using geometric methods for route searching \cite{harabor2011online,liu2017planning} and stitched polynomial formulation for trajectory optimization \cite{mellinger2011minimum,richter2016polynomial}. However, these methods do not consider the quadrotor's dynamic model in the route searching and optimization process, and the trajectory optimized with geometric methods that may exceed the dynamic limitations of the quadrotor. 
Generating and tracking feasible trajectories in dynamic environments is typically achieved by directly feeding any aerodynamic disturbances into the tracking module and controller. Such an approach does not consider the entire quadrotor system framework - e.g., route searching, trajectory optimization and tracking, and the flight controller - and leaves the quadrotor vulnerable to attempting to navigate unsuitable environments.

\begin{figure}[]
  \centering
  \includegraphics[scale=0.38]{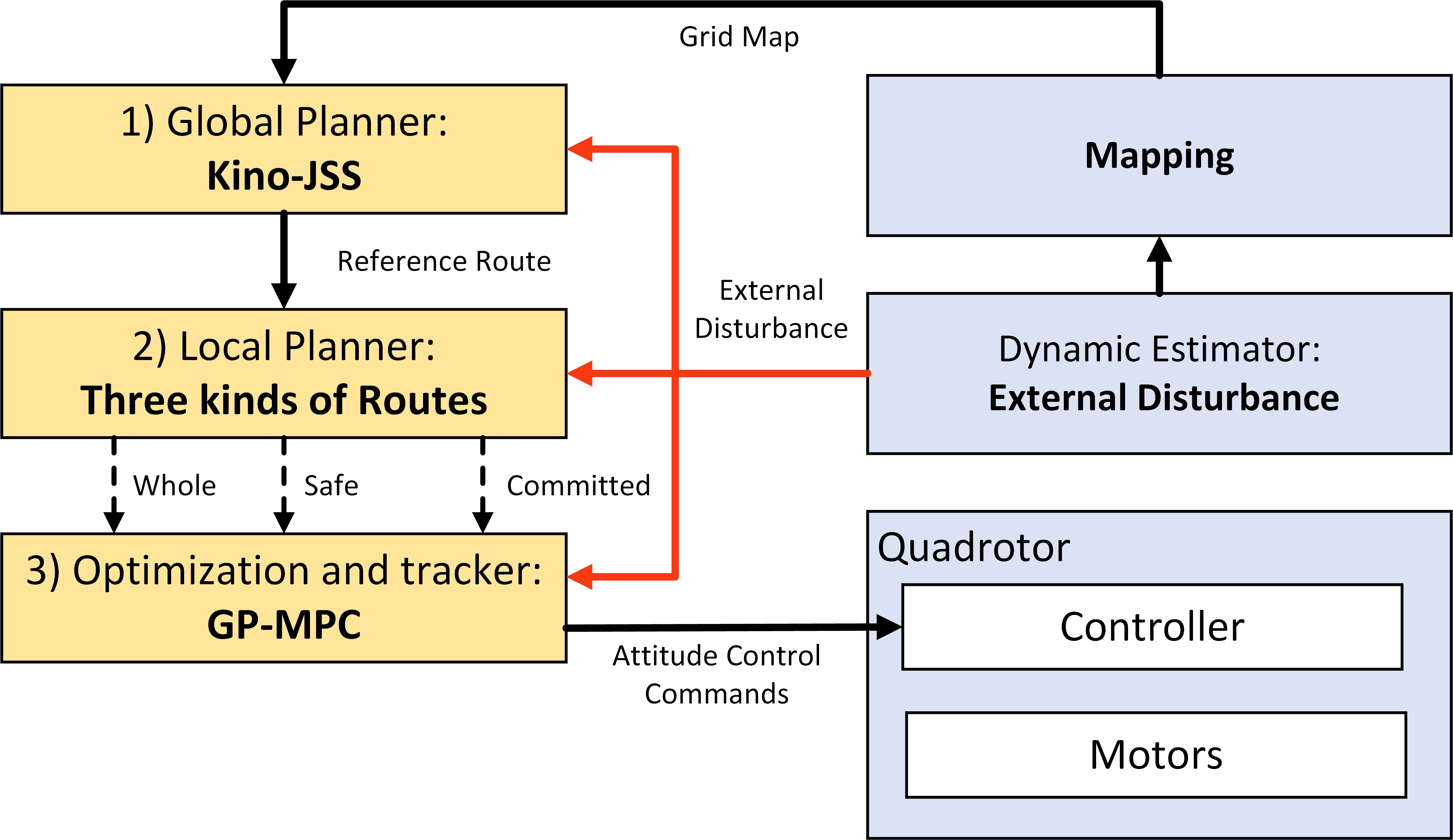}
  \caption{System overview of the KinoJGM framework: given the aerodynamic estimation, Kino-JSS (global planner), local planner and GP-MPC (optimization and tracker) generate trajectories at a high rate and give the attitude control commands to the quadrotor's controller in real-time.}
  \label{system_diagram}
\end{figure}

\begin{figure}[]
  \centering
  \includegraphics[scale=0.87]{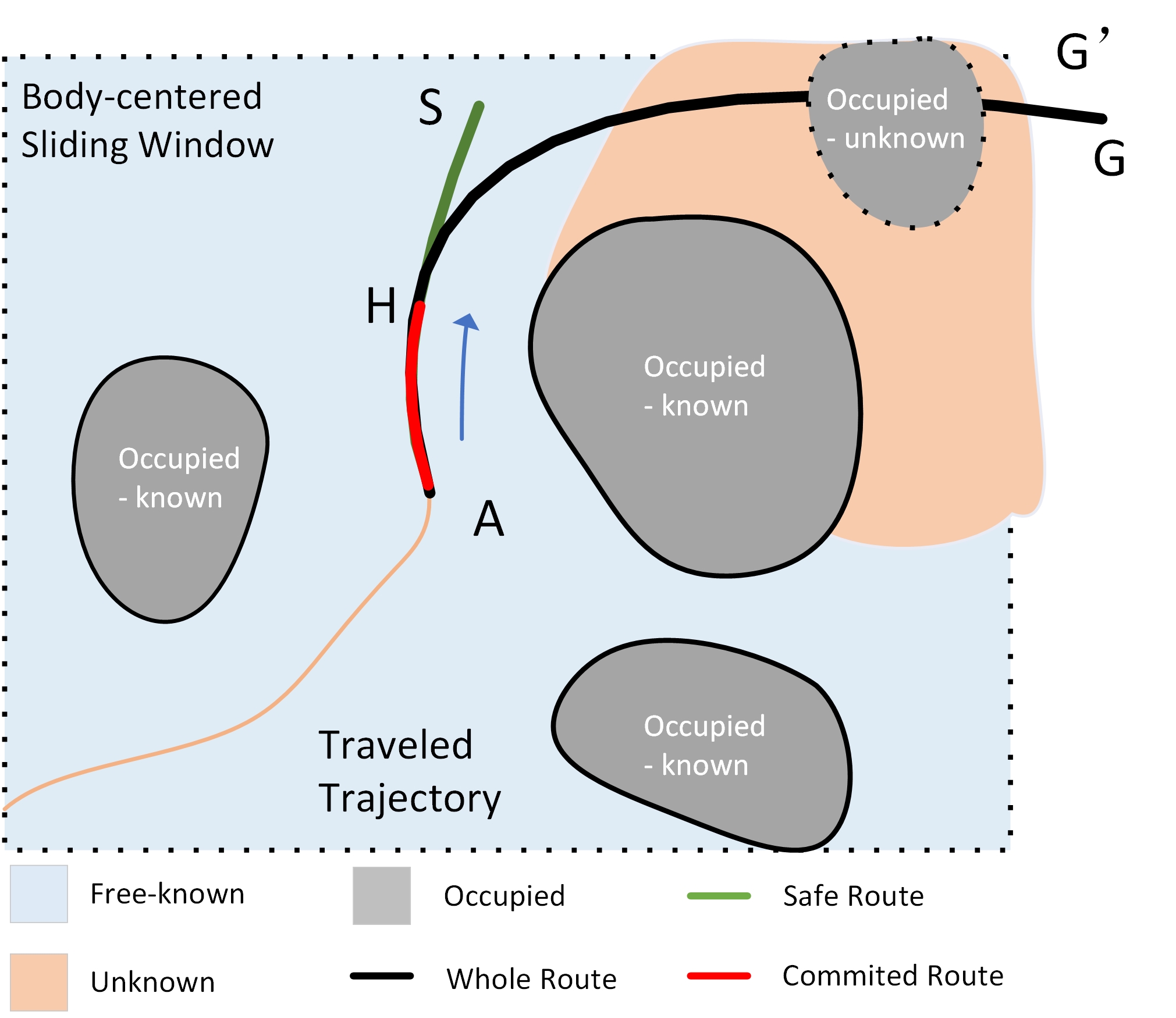}
  \caption{Local planner: whole, safe and committed routes are generated, respectively. Point A is the current position of quadrotor, G' is the next goal position, and, G is the terminal goal position.}
  \label{local_planner}
\end{figure}

To address the two stated issues, we propose \textbf{Kino}dynamic \textbf{J}ump Space Search and \textbf{G}aussian-Process-based \textbf{M}odel Predictive Control (KinoJGM), a systematic, safe and feasible online trajectory generation and tracking framework for use in unknown environments with aerodynamic disturbances (shown in Fig. \ref {system_diagram}).
The procedure is as follows:
\begin{enumerate}
    \item Global Planner: a Kinodynamic Jump Space Search (Kino-JSS), described by Algorithms \autoref{Kino_JSS},  \autoref{KinoJSSRecursion} and \autoref{JSSMotion}, is proposed for route searching. Kino-JSS builds upon traditional Jump Point Search (JPS) \cite{harabor2011online,liu2017planning} insofar as Kino-JSS is a kinodynamic process which considers the dynamic model of the quadrotor whilst generating routes. Kino-JSS runs an order of magnitude faster than a kinodynamic search based on a hybrid-state A* search \cite{zhou2019robust}, whilst maintaining comparable performance. Kino-JSS can not only generate a safe and feasible original route efficiently, but can also make adjustments based on aerodynamic disturbance estimation and the quadrotor dynamic limitation. 
    \item Local Planner: Similar to \cite{tordesill2021faster}, three types of route are distinguished in real-time based on the original route points given by the Kino-JSS. As shown in Figure \ref {local_planner}, the whole route is from $A$ to $G'$, which is efficient and has an end condition; the safe route is from $A$ to $S$; and the committed route is from $A$ to $H$, which is safe and efficient. 
    \item Optimization and tracker: an integrated Model Predictive Control (MPC) using Gaussian Process (GP) is proposed to optimize routes from the Local Planner and track accurately. In order to decrease the computational cost, the integrated MPC is constrained into a corresponding polyhedron \cite{liu2017planning} derived from the Local Planner routes.
\end{enumerate}

Our contributions can be summarized as follows:
\begin{itemize}
\item[1)]
Kino-JSS, a route searching algorithm with aerodynamic disturbance estimation that can efficiently generate safe and feasible routes.
\end{itemize}
\begin{itemize}
\item[2)]
The integration of aerodynamic disturbance estimation with GP-MPC, a real-time trajectory optimization and tracking system that integrates GP into an optimal control problem constrained in polyhedrons.
\end{itemize}
\begin{itemize}
\item[3)]
KinoJGM, a systematic trajectory planning and tracking framework verified in simulated experiments.
\end{itemize}

\section{Related Work}
The literature on quadrotor trajectory generation and tracking is extensive and comprises a wide variety of backgrounds and perspectives - e.g., search-based methods \cite{liu2017search,liu2018search,aine2016multi}, sampling-based methods \cite{webb2013kinodynamic,janson2015fast,gammell2015batch,allen2016real}, optimization-based methods \cite{gao2016online,chen2016online,oleynikova2016continuous} and control-based methods \cite{van2012lqg,bareiss2015stochastic}. For brevity, we use this section to discuss the most relevant literature, which we arrange into two categories: kinodynamic trajectory planning, and robust planning with dynamic disturbance.

\subsection{Kinodynamic trajectory planning}
Kinodynamic trajectory planning explores optimized trajectories in a high-dimensional state space and outputs a time-parameterized trajectory \cite{ding2019efficient}, in which quadrotor dynamics are considered. These methods - e.g., \cite{webb2013kinodynamic,janson2015fast,gammell2015batch,lavalle2001randomized} - correspond to kinematic systems which provide an efficient and easy way to achieve kinodynamic searching. An efficient kinodynamic Boundary Value Problem solver is designed by Xie et al. \cite{xie2015toward} using sequential quadratic programming. Although the efficiency of kinodynamic searching keeps improving, it remains a computationally expensive process, which limits its suitability for online planning. The work by Liu et al. \cite{liu2017search} and Allen et al. \cite{allen2016real} towards a real-time kinodynamic planning framework develops efficient heuristics by solving a linear quadratic minimum time problem. However, the generated trajectory is not always smooth as they use a simplified system model. Ding et al. \cite{ding2019efficient,ding2018trajectory} propose a kinodynamic search with B-spline optimization. The uniform B-spline improves feasibility and safety for the quadrotor, but the use of non-adaptive time allocation polyhedrons can significantly reduce overall trajectory quality in some scenarios, and adding more constraints iteratively is prohibitive for real-time applications. Zhou et al. \cite{zhou2019robust} propose a robust Kinodynamic-Route-Search (Kino-RS) method in discretized control space. This online kinodynamic method is robust and provides dynamic feasibility. Building on this work \cite{zhou2019robust}, our simulated experiments demonstrate that Kino-JSS runs an order of magnitude faster than Kino-RS \cite{zhou2019robust} in obstacle-dense environments, whilst maintaining comparable system performance.

\subsection{Robust Planning with dynamic disturbance}
Many environments require an autonomous quadrotor system to resist aerodynamic disturbances whilst planning a safe and energy-efficient path. Typically, dynamic disturbances are handled by the quadrotor's tracking module and flight controller \cite{yu2012sliding,liu2008adaptive}, because robust control is generally designed for the “worst-case” bounded uncertainties and disturbances \cite{zhang2020model}. Adaptive controllers \cite{aastrom2013adaptive} can address some of the problems caused by aerodynamic disturbances with unknown boundaries, but the system needs to be continuously updated with vehicle state estimations for effective control. For a quadrotor, this can cause undesired high-frequency oscillation behaviours. Singh et al. \cite{singh2017robust}, Guerrero \cite{guerrero2013quad} and Palmieri \cite{palmieri2017kinodynamic} design a global planner for a simulated quadrotor flying in areas with aerodynamic disturbances. However, this method is tested in highly controlled simulated environments, and assumes accurate aerodynamic disturbance estimation.

MPC is an efficient tool to generate optimal local trajectories and encode uncertainty arising from aerodynamic disturbances. In \cite{mehndiratta2020gaussian}, GP is used on a quadrotor to correct for wind disturbances. Hewing et al. \cite{hewing2019cautious}, Kabzan et al.~\cite{kabzan2019learning} and Torrente et al.~\cite{torrente2021data} design and implement a GP-MPC method to accurately predict small aerodynamic disturbances and improve tracking accuracy for autonomous vehicles. Building on these works \cite{hewing2019cautious,kabzan2019learning,torrente2021data}, we integrate GP and MPC, but extend to large aerodynamic disturbances - e.g., strong winds and those generated by the quadrotor and environmental obstacles. Our proposed GP-MPC sets the estimation of aerodynamic disturbance as a value in formulation (\autoref{state_pro} and \autoref{state_pro_with_noise}) to generate optimized trajectories and perform accurate tracking in real-time.

\section{Methodology}
In this section, we present the proposed KinoJGM framework and its novel modules. This work addresses the current limitations on route searching and trajectory planning in unknown environments with aerodynamic disturbances. The modules that comprise our KinoJGM framework are described as follows.

\subsection{Mapping Module}
In practice, drone flight is oftentimes desirable in GPS-denied environments, where on-board sensing will be relied on for state estimations -  e.g., position, velocity, acceleration and jerk. The accumulation of drift is therefore a significant problem when operating in complex and/or large environments. We use a body-centered sliding window, as described in \cite{tordesill2021faster}, \cite{schmid2021unified}, and shown in \autoref{local_planner}, to reduce the influence of drift in state estimation error.

\subsection{Aerodynamic Disturbance Estimation Module}
We use Visual-Inertial-Dynamics-Fusion (VID-Fusion) \cite{ding2020vid} to estimate the aerodynamic disturbance operating on the quadrotor at a given time. VID-Fusion is a tightly coupled  state estimator for quadrotors. It compares the quadrotor dynamic model with the inertial measurement unit (IMU) reading - i.e., Visual-Inertial-System-Monocular (VINS-mono) \cite{qin2018vins} and Visual Inertial Model-Based Odometry (VIMO) \cite{nisar2019vimo} - to generate robust and accurate aerodynamic disturbance and pose estimation.

\subsection{Route Searching Module (Kino-JSS)}
Quadrotor route searching primarily focuses on robustness, feasibility and efficiency. The Kino-RS algorithm \cite{zhou2019robust} is a robust and feasible online searching approach. However, the searching loop is derived from the hybrid-state A* algorithm, making it relatively inefficient in obstacle-dense environments. On the other hand, JPS offers robust route searching, and runs at an order of magnitude faster than the A* algorithm \cite{harabor2011online}. A common problem of geometric methods such as JPS and A* is that, unlike kinodynamic searching, they consider heuristic cost (e.g., distance) but not the quadrotor dynamics and feasibility (e.g., line 5 of Algorithm \autoref{KinoJSSRecursion} and line 10 of Algorithm \autoref{JSSMotion}) when generating routes \cite{ding2019efficient}. $\bf{checkFea()}$ is the feasibility check to judge the acceleration and velocity constrains based on the quadrotor dynamics. Kino-JSS is described by Algorithms \autoref{Kino_JSS}, \autoref{KinoJSSRecursion} and \autoref{JSSMotion}.

\begin{algorithm}[]
\caption{Kinodynamic Jump Space Search}
\bf{INPUT}: $s_{cur}$\\
\bf{OUTPUT}: $KinoJSSRoute$\
\begin{algorithmic}[1]
\STATE $\bf{initialize()}$
\STATE $openSet.\bf{insert(\it{s_{cur}})}$
\WHILE {!$openSet.\bf{isEmpty()}$}
\STATE $s_{cur} \gets openSet.\bf{pop()}$
\STATE $closeSet.\bf{insert(\it{s_{cur}})}$
\IF{$\bf{nearGoal(\it{s_{cur}})}$}
\RETURN $KinoJSSRoute$
\ENDIF
\STATE $\bf{KinoJSSRecursion()}$
\ENDWHILE 
\end{algorithmic}
\label{Kino_JSS}
\end{algorithm}

\begin{algorithm}[]
\caption{KinoJSSRecursion}
\bf{INPUT}: $s_{cur}$, $E_f$, $openSet$, $closeSet$\\
\bf{OUTPUT}: $void$\
\begin{algorithmic}[1]
\STATE $motions \gets \bf{JSSMotion(\it{s_{cur}, E_f})}$
\FOR {each $m_i \in motions$}
\STATE $s_{pro} \gets \bf{statePropagation(\it{s_{cur},m_i}))}$
\STATE $inClose \gets \it{closeSet}.\bf{isContain(\it{s_{pro}})}$
\IF{$\bf{isFree(\it{s_{pro}})} \bigwedge \bf{checkFea(\it{s_{pro},m_i})} \bigwedge \it{inClose}$}
\IF{ $\bf{checkOccupiedAround(\it{s_{pro}})}$}
\STATE $s_{pro}.neighbors \gets \bf{JSSNeighbor(\it{s_{pro}})}$
\STATE $cost_{pro} \gets \it{s_{cur}}.cost+ \bf{edgeCost(\it{s_{pro}})}$
\STATE $cost_{pro} \gets cost_{pro} + \bf{heuristic(\it{s_{pro}})}$
\IF{!$\it{openSet}.\bf{isContain(\it{s_{pro}})}$}
\STATE $openSet.\bf{insert(\it{s_{pro}})}$
\ELSIF{$\it{s_{pro}}.cost\leq \it{cost_{pro}}$}
\STATE continue
\ENDIF
\STATE $s_{pro}.parent \gets s_{cur}$
\STATE $s_{pro}.cost \gets cost_{pro}$
\ELSE
\STATE $\bf{KinoJSSRecursion()}$
\ENDIF
\ELSE
\STATE continue
\ENDIF
\ENDFOR
\end{algorithmic}
\label{KinoJSSRecursion}
\end{algorithm}

\begin{algorithm}[]
\caption{JSSMotion}
\bf{INPUT}: $s_{cur}$, $E_f$\\
\bf{OUTPUT}: $motions$\
\begin{algorithmic}[1]
\STATE $E_{fcor} \gets E_f + \bf{GaussianNoise()}$
\FOR{each $m_i \in motionSet$}
\STATE $m_{cor} \gets m_i + E_{fcor}$
\STATE $motions \gets \bf{push\_back(\it{m_{cor}})}$
\ENDFOR
\STATE $neighSize \gets  s_{cur}.neighbors.\bf{size()}$
\WHILE{$neighSize \neq 0$}
\STATE $neighSize = neighSize - 1$
\STATE $neighMotion \gets \bf{posToMotion(\it{s_{cur}.neighbors)}}$
\IF{$\bf{checkFea(\it{s_{cur},neighMotion})}$}
\STATE $motions \gets \bf{push\_back(\it{neighMotion})}$
\ENDIF
\ENDWHILE
\RETURN $motions$
\end{algorithmic}
\label{JSSMotion}
\end{algorithm}

\begin{figure}[t]
  \centering
  \includegraphics[scale=0.87]{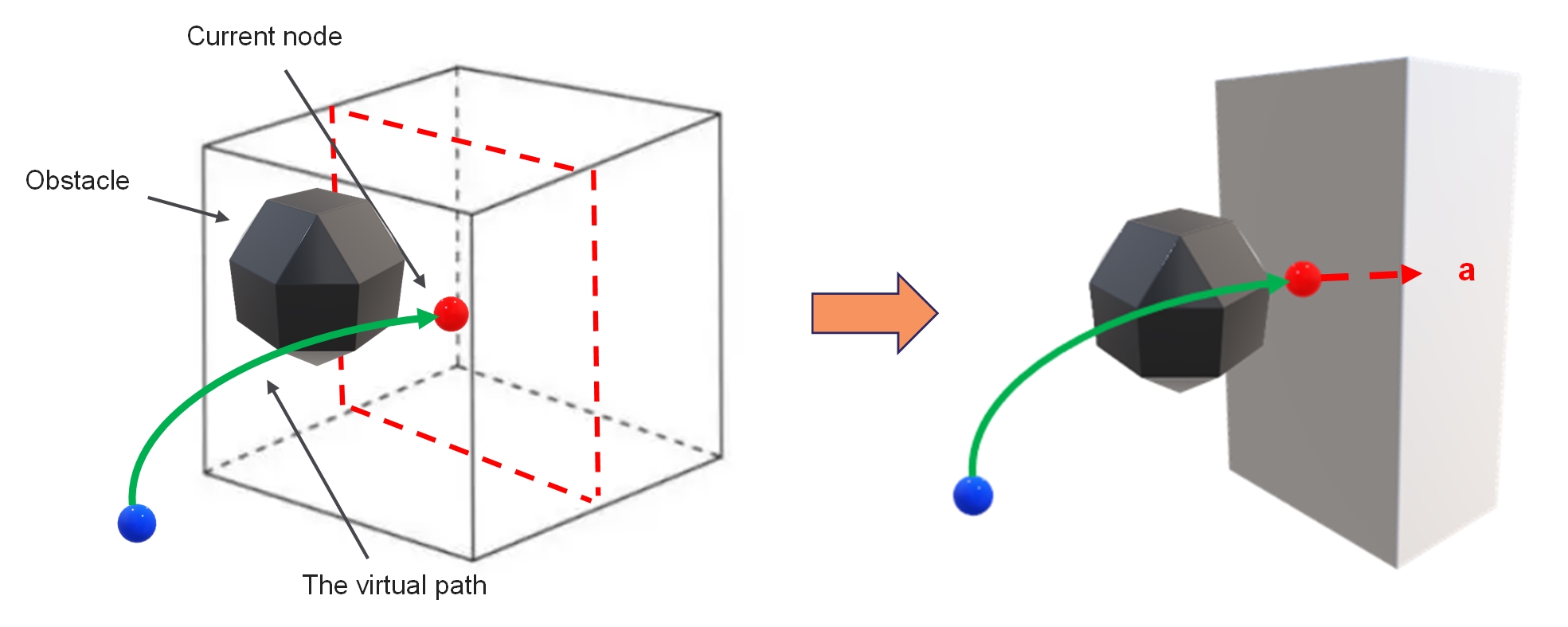}
  \caption{Diagram of KinoJSS when checking occupied obstacles around: a grey cuboid is generated as the ‘\textbf{forced neighbour}’, the output of $\bf{JSSNeighbor()}$, in Algorithm~2.}
  \label{occumotion}
\end{figure}

\begin{figure}[t]
  \centering
  \includegraphics[scale=0.87]{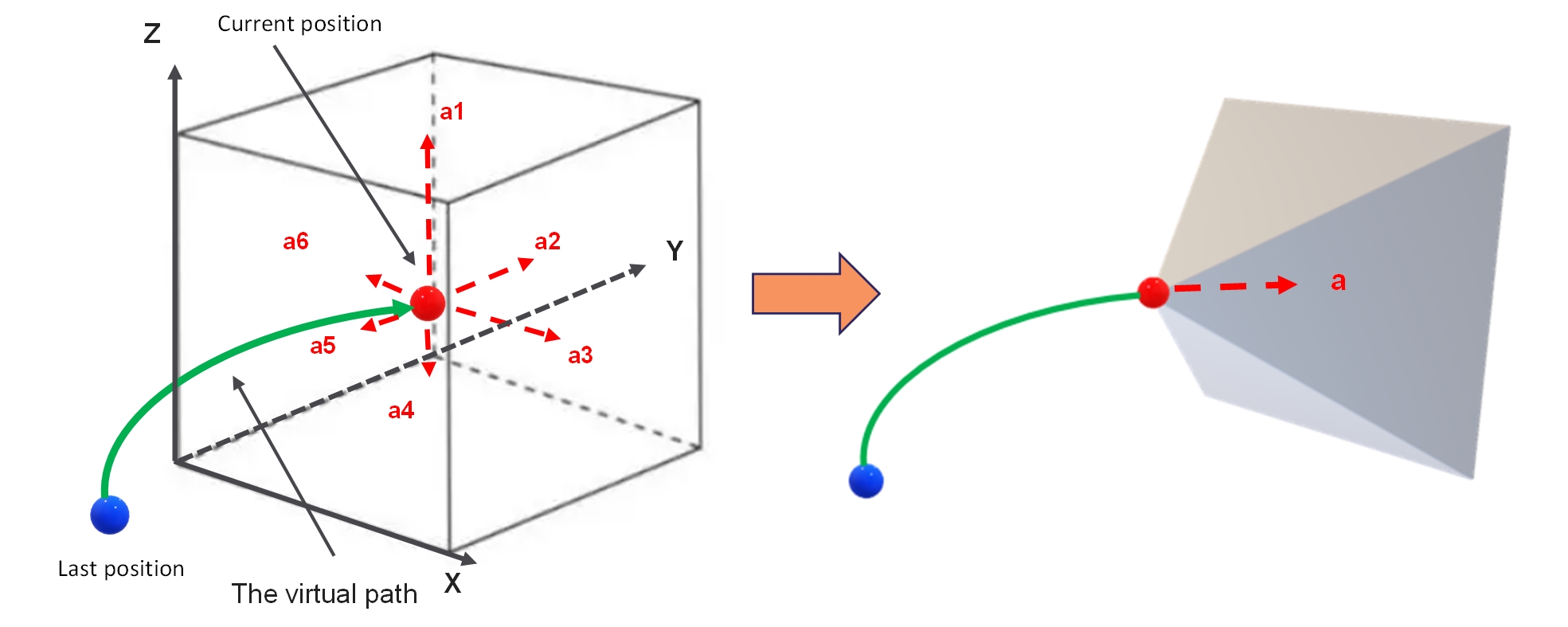}
  \caption{Diagram of KinoJSS with unoccupied space around: the pyramid represents the $motionSet$ in Algorithm~3, i.e., $\bf{JSSMotion()}$, in free space.}
  \label{freemotion}
\end{figure}

$s_{cur}$ denotes the current state, $s_{pro}$ denotes the propagation of current state under the motion $m_i$, and $E_f$ denotes the aerodynamic disturbance estimated by VID-fusion \cite{ding2020vid}. In Algorithm~2, $\bf{JSSNeighbor()}$ is defined as shown in Fig.~\ref{occumotion}. In Algorithm~3, $motionSet$, which is defined as a pyramid shown in Fig.~\ref{freemotion}, offers improved efficiency whilst retaining the advantages of Kino-RS \cite{zhou2019robust}.

\subsection{Trajectory Optimization and Tracking Module (GP-MPC)}
Traditional approaches \cite{tordesill2021faster,liu2017planning} address trajectory optimization and tracking separately. This partitioned solution seems to be logical and each process is easy to analyze in isolation. On the other hand, integrating the trajectory optimization and tracking processes increases the search space and computing complexity, which limits the quadrotor's ability to react dynamically in real-time \cite{neunert2016fast}.
We use an integrated MPC-based approach that generates an optimized trajectory and a sequence of tracking commands by solving an optimal control problem. Trajectory optimization and tracking is achieved using GP-MPC, which combines with a full-state quadrotor MPC formulation \cite{mueller2015computationally,falanga2018pampc,torrente2021data}.

\begin{figure}[]
  \centering
  \includegraphics[scale=0.6]{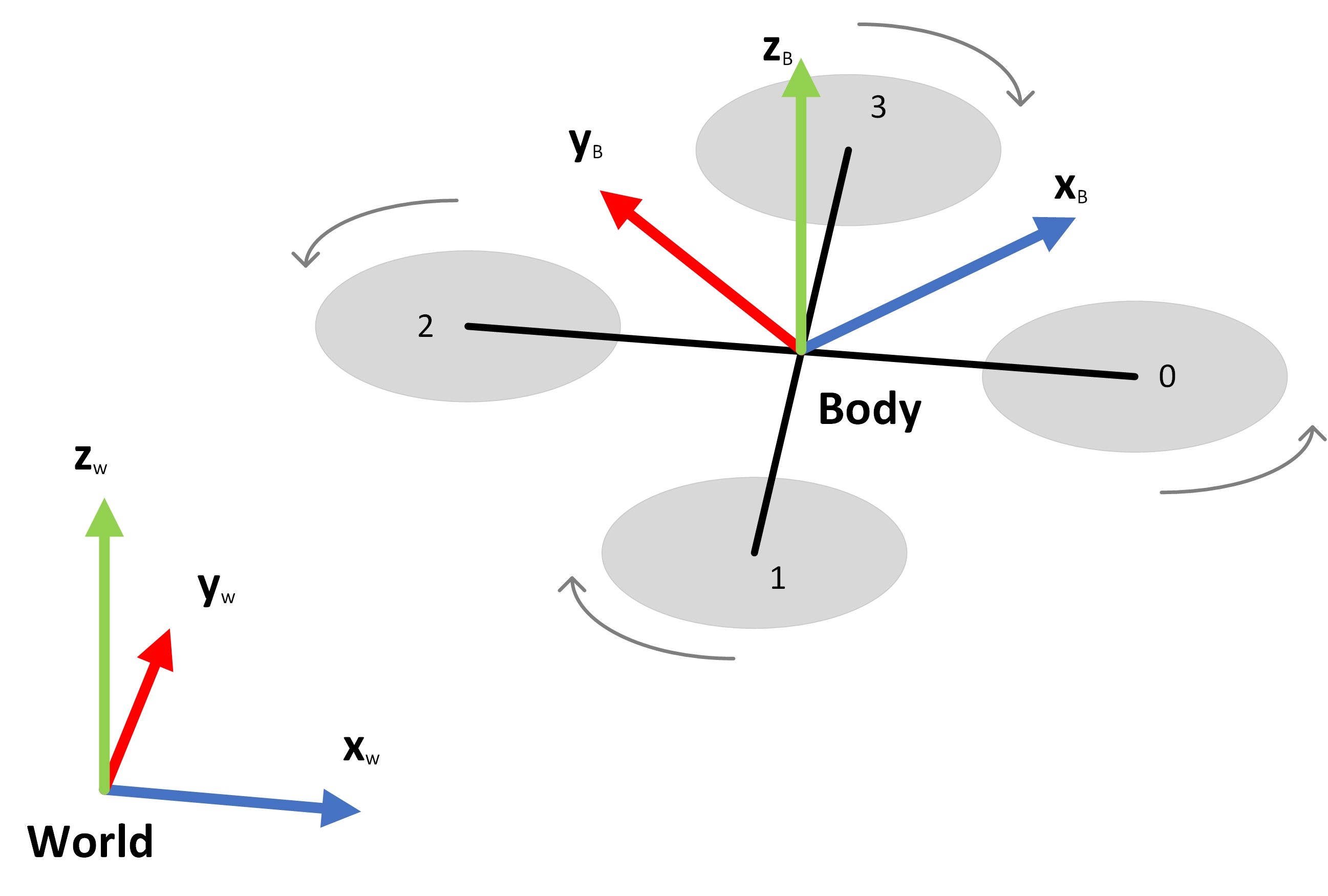}
  \caption{Diagram of the quadrotor the world frame W and the body frame B.}
  \label{world_frame_body_frame}
\end{figure}

We integrate the aerodynamic force (disturbance) $\bm{e}_f$ into the dynamic model of a quadrotor as follows:
\begin{equation}
\begin{aligned}
&\dot{\bm{P}}_{WB} = \bm{V}_{WB}\\
&\dot{\bm{V}}_{WB} = \bm{g}_W + \frac{1}{m}(\bm{q}_{WB}\odot \bm{c} + \bm{e}_f)\\
&\dot{\bm{q}}_{WB} = \frac{1}{2}\Lambda(\bm{\omega} _{B})\bm{q}_{WB}\\
&\dot{\bm{\omega}} _{B} = \bm{J}^{-1}(\bm{\tau}_B-\bm{\omega}\times J\bm{\omega} _{B})
\end{aligned}
\label{quadrotor_dynamics}
\end{equation}
where $\bm {P}_{WB}$, $\bm{V}_{WB}$ and $\bm{q}_{WB}$ are the position, linear velocity and orientation of the quadrotor expressed in the world frame (shown in  Fig.~\ref{world_frame_body_frame}), and $\bm{\omega}_{B}$ is the angular velocity expressed in the body frame \cite{torrente2021data}. The state of the dynamic model - i.e., $\bm{x} = [\bm{P}_{WB},\bm{V}_{WB},\bm{q}_{WB},\bm{\omega} _{B}]^T$ - is used for quadrotor motion planning. The input vector of the dynamic model is $\bm{u} = T_i \forall i\in(0,3)$. $\bm{c}$ is the collective thrust $\bm{c} = [0,0,\sum T_i]^T$ and $\bm{\tau}_B$ is the body torque; $\bm{g}_W = [0,0,-g]^T$. The operator $\odot$ denotes a rotation of the vector by the quaternion. We assume that the aerodynamic force $\bm{e}_f$ in \autoref{quadrotor_dynamics} only effects the position motion of quadrotor and not its angular motion. The skewsymmetric matrix $\Lambda(\bm{\omega})$ is defined in \cite{falanga2018pampc}.


For integration in a discrete-time MPC, the quadrotor dynamic model is discretized using a Runge-Kutta 4th-order integration \cite{hewing2019cautious} with sampling time $T_s = 50ms$. As per \cite{ding2020vid}, the nominal force $\bm{n_f}$ can be treated as a constant value calculated in short duration, and the aerodynamic force estimation is defined as: $\bm{e}_{fk} = \bm{n_f} + \bm{w}_k$, where $\bm{w}_k \sim \mathcal N(0, \sigma)$ is Gaussian noise. So, the discrete-time state $\bm{x}_k$ is defined as:
\begin{equation}
\bm{x}_{k+1} = \bm{f}_{RK}(\bm{x}_{k},\bm{u}_{k}, \bm{e_{fk}})
\label{state_pro}
\end{equation}

We define the control of the quadrotor dynamic model with a discrete-time model:
\begin{equation}
\begin{aligned}
\bm{x}_{k+1} = \bm{f}(\bm{x}_{k},\bm{u}_{k}, \bm{e_{fk}}) + \bm{B}_d(\bm{g}(\bm{x}_{k},\bm{u}_{k})+\bm{w}_k)
\label{state_pro_with_noise}
\end{aligned}
\end{equation}
where $B_d$ is the selection matrix, $\bm{g}$ describes an initially unknown dynamic model, and, both $\bm{f}$ and $\bm{g}$ are assumed to be differentiable functions. 

In GP-MPC, GP is used to predict the error of the dynamic model. Like most GP-based approaches, e.g.,~\cite{hewing2019cautious,kabzan2019learning,torrente2021data}, we formulate this by assuming a true value $\bm{f}_{true}$ and a measurement value $\Tilde{\bm{x}}_{k+1} = \bm{f}_{mea}$:
\begin{equation}
\begin{aligned}
\Tilde{\bm{x}}_{k+1} &= \bm{f}_{mea}(\bm{x}_{k},\bm{u}_{k},\bm{e_{fk}}) \\
&= \bm{f}_{true}(\bm{x}_{k},\bm{u}_{k},\bm{e_{fk}}) + \bm{w}_k
\end{aligned}
\end{equation}
where $\bm{w}_k \sim \mathcal N(0, \sigma)$ is the process noise and $\sum$ is a diagonal variance matrix.

We use the rational quadratic kernel as a kernel function $\bm{k}(\bm{z}_i,\bm{z}_j)$ which is defined by:
\begin{equation}
\begin{aligned}
\bm{k}(\bm{z}_i,\bm{z}_j) = \sigma_{f}^{2}(1+\frac{(\bm{z}_i-\bm{z}_j)^{T} L^{-2}(\bm{z}_i-\bm{z}_j)}{2\alpha})^{-\alpha}+\sigma_{n}^{2}
\end{aligned}
\end{equation}
where $L$ is the diagonal length scale matrix; $\sigma_{f}$ and $\sigma_{n}$ represent the data and prior noise variance; and, $\bm{z}_i$,$\bm{z}_j$
represents data features. Then, the mean and covariance of the GP is defined as follows:
\begin{equation}
\begin{aligned}
\bm{\mu}(Z) =& (\bm{k}(Z,Z_k))^{T}(\bm{k}(Z,Z)+\sigma_{n}^{2}I)^{-1}Z\\
\Sigma(Z) =& \bm{k}(Z_k,Z_k) \\
&- (\bm{k}(Z,Z_k))^{T}(\bm{k}(Z,Z)+\sigma_{n}^{2}I)^{-1}\bm{k}(Z,Z_k)
\end{aligned}
\end{equation}
where $Z$ and $Z_k$ denote the training input and test samples.
Given the mean and covariance of the GP, we calculate the modified dynamic and feed it into the MPC formulation as follows:
\begin{equation}
\begin{aligned}
\mathop{min}\limits_{u} L(\bm{x},\bm{u}) = &\bm{x}_N^TQ\bm{x}_N+\sum_{k=0}^N\bm{x}_k^TQ\bm{x}_k+\bm{u}_N^TR\bm{u}_N\\
\rm{subject}\quad \rm{to}\quad &\bm{x}_{k+1} = \bm{f}(\bm{x_k},\bm{u_k},\bm{e_{fk}})\\
&\bm{x_0} = \bm{g}(\bm{x_k},\bm{u_k})\\
&\bm{u_{min}}\leq\bm{u}\leq\bm{u_{max}}
\end{aligned}
\label{mpcformulation}
\end{equation}

The quadratic optimization formulation in \autoref{mpcformulation} is discretized based on the multiple shooting method \cite{diehl2006fast} and solved using sequential quadratic programming for real-time motion planning \cite{diehl2006fast}. We implement the calculation of the MPC formulation using CasADi \cite{andersson2019casadi} and ACADOS \cite{verschueren2018towards}.
\begin{figure}[]
  \centering
  \includegraphics[scale=0.42]{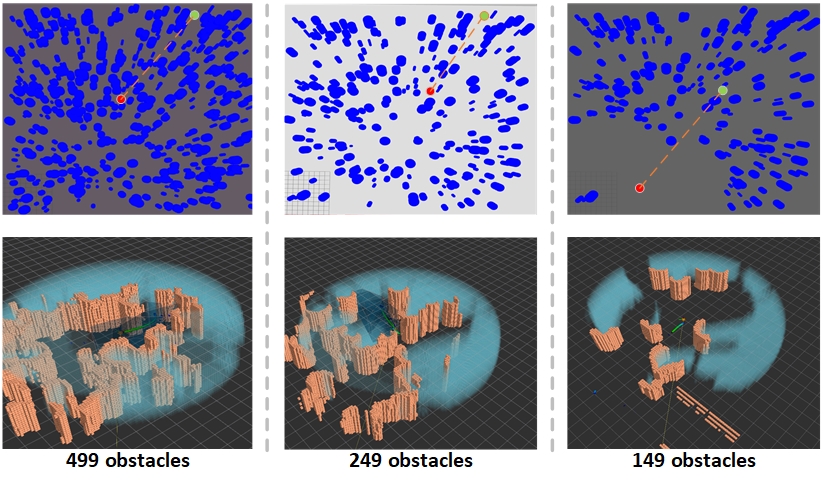}
  \caption{Three kinds of environment with different obstacles (499, 249 and 149 obstacles) in the simulation.}
  \label{environment}
\end{figure}

\section{Implementation and Results}
\subsection{Setup}
We evaluate the performance of our proposed trajectory generation and tracking framework in software simulation, in which a DJI Manifold 2-C (Intel i7-8550U CPU) is used for real-time computation. We use RotorS MAVs simulator \cite{furrer2016rotors}, in which programmable aerodynamic disturbances can be generated, for testing the KinoJGM framework. As shown in Section \uppercase\expandafter{\romannumeral3}-D, the aerodynamic disturbance is defined and simulated as a nominal force $\bm{n_f}$ plus Gaussian noise $\bm{w}_k$. The nominal force $\bm{n_f}$ is estimated by VID-Fusion \cite{ding2020vid}. The noise bound of aerodynamic disturbance is set as 0.5 $m/s^2$, based on the benchmark established in \cite{wu2021external}. We first test the Kino-JSS (route searching) module against the system described in \cite{zhou2019robust}, before comparing our entire KinoJGM framework with the system described in \cite{wu2021external}. Since the update frequency of the aerodynamic force $\bm{e_f}$ estimation is much higher than our KinoJGM framework frequency, we sample $\bm{e_f}$ based on our framework frequency. We also assume the collective thrust $\bm{c}$ is a true value, which is tracked ideally in the simulation platform. 



\subsection{Experiments}

\begin{table*}[]
\caption{Comparison of Route Searching}
\label{Comparison_of_route_searching}
\begin{center}
\begin{tabular}{|c|l|c|c|c|c|c|c|c|c|c|}
\hline
\multicolumn{2}{|c|}{\multirow{2}{*}{}} & \multicolumn{3}{c|}{499 Obstacles}                & \multicolumn{3}{c|}{249 Obstacles}                 & \multicolumn{3}{c|}{149 Obstacles}                 \\ \cline{3-11} 
\multicolumn{2}{|c|}{}                  & Time (ms) & Ctrl cost & Succ. Rate          & Time (ms) & Ctrl cost & Succ. Rate           & Time (ms) & Ctrl cost & Succ. Rate           \\ \hline
\multirow{3}{*}{Zhou. \cite{zhou2019robust}}       & Mean    & 70.3664    & 57.73      & \multirow{3}{*}{78.60\%} & 39.6634    & 46.11      & \multirow{3}{*}{100.00\%} & 23.6290    & 21.25      & \multirow{3}{*}{100.00\%} \\ \cline{2-4} \cline{6-7} \cline{9-10}
                              & Std     & 55.5857    & 29.78      &                          & 32.7760    & 28.94      &                           & 13.7430    & 36.78      &                           \\ \cline{2-4} \cline{6-7} \cline{9-10}
                              & Max     & 236        & 101.18      &                          & 232        & 97.25      &                           & 75        & 98.31      &                           \\ \hline
\multirow{3}{*}{Kino-JSS}     & Mean    & \bf{17.3782}   & \bf{48.12}      & \multirow{3}{*}{79.61\%} & \bf{13.1701}    & \bf{28.75}      & \multirow{3}{*}{100.00\%} & \bf{10.9858}     & \bf{19.59}      & \multirow{3}{*}{100.00\%} \\ \cline{2-4} \cline{6-7} \cline{9-10}
                              & Std     & 10.8848     & 11.30      &                          & 7.5069     & 15.56      &                           & 5.8236      & 27.36      &                           \\ \cline{2-4} \cline{6-7} \cline{9-10}
                              & Max     & 106         & 72.06      &                          & 39         & 61.02      &                           & 34         & 58.38      &                           \\ \hline
\end{tabular}
\end{center}
\end{table*}
1) Comparative performance of Kino-JSS route searching module: We compare our Kino-JSS module with Zhou et al. \cite{zhou2019robust}, which uses a Kino-RS method to generate route points. The comparison is on a $40m\times40m\times5m$ map, where 1m of simulated space corresponds to 0.15m of real space. The maximum quadrotor velocity and acceleration are set as $3m/s$ and $2m/s^2$, respectively (in real space). These parameters are equivalent to those used by Zhou et al. \cite{zhou2019robust}. As shown in \autoref{environment}, we perform experiments in simulated environments with 499, 249 and 149 obstacles to demonstrate our system's performance in environments of varying obstacle-density.

Unlike Kino-RS \cite{zhou2019robust}, Kino-JSS will not add a search point to the $openSet$ until an obstacle is encountered. This characteristic of Kino-JSS means that more space can be searched per operation, and substantially increases the efficiency of Kino-JSS over Kino-RS in obstacle-dense environments. This is demonstrated by our results given in \autoref{Comparison_of_route_searching}. The improved efficiency of Kino-JSS is most evident in areas with highest obstacle density, where we achieve improvements of 75\%, 67\% and 53\% for environments with 499, 249 and 149 obstacles, respectively. Predictably, the improvement in efficiency for Kino-JSS decreases with lower obstacle density, since it is proved in \cite{harabor2011online} that the performance of a JPS algorithm is equivalent with an A* algorithm in free environments.
Our results also show that the quadrotor control cost of Kino-JSS is much lower than Kino-RS, because Kino-JSS can jump way-points and only operates way-points with forced neighbours, also shown in line 7 of Algorithm \autoref{KinoJSSRecursion} and line 1 of Algorithm \autoref{JSSMotion}.

\begin{figure}[]
  \centering
  \includegraphics[scale=0.57]{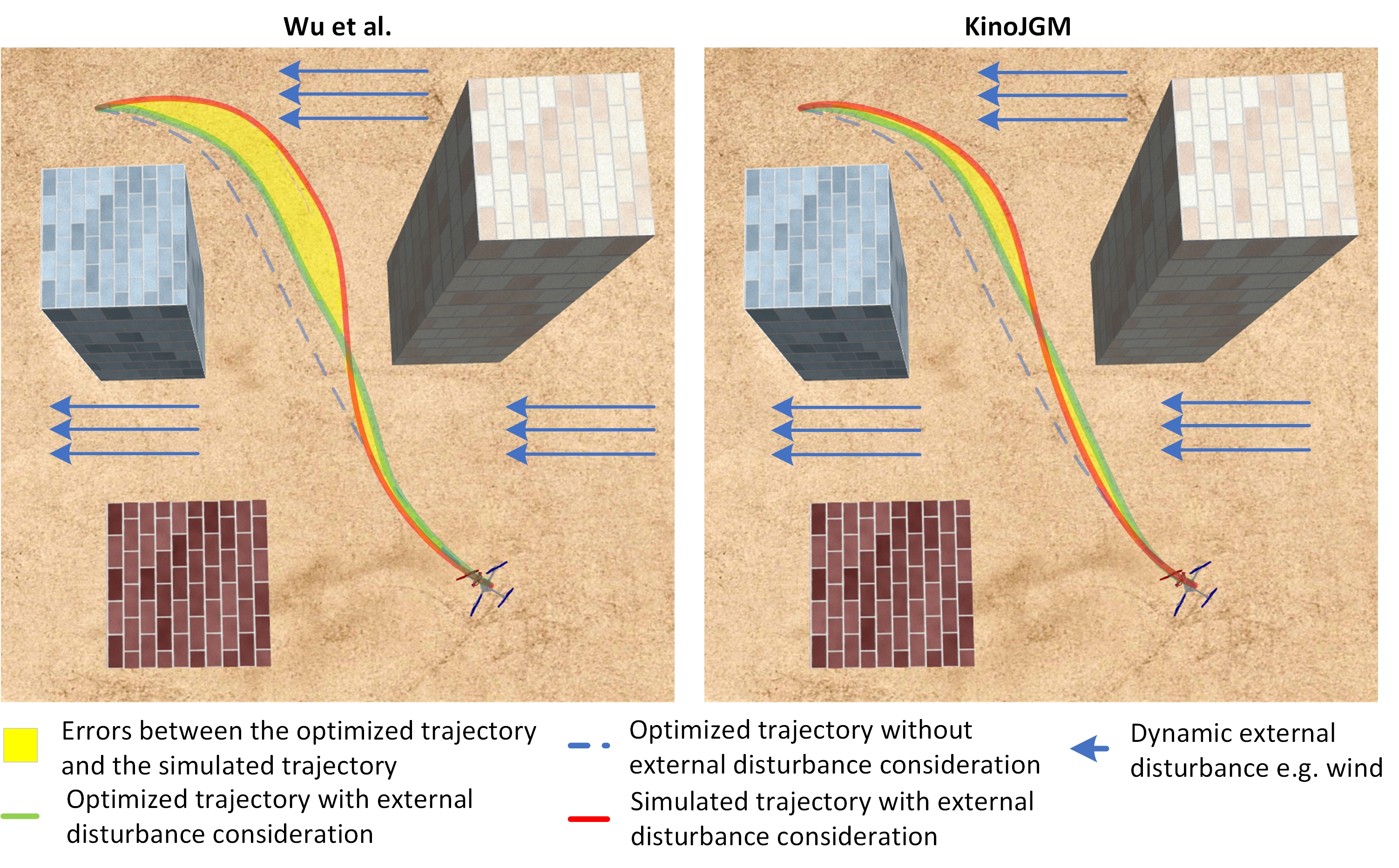}
  \caption{The comparison between Wu et al. \cite{wu2021external} and our proposed KinoJGM framework.}
  \label{simulation_env}
\end{figure}

\begin{figure}[]
  \centering
  \includegraphics[scale=0.205]{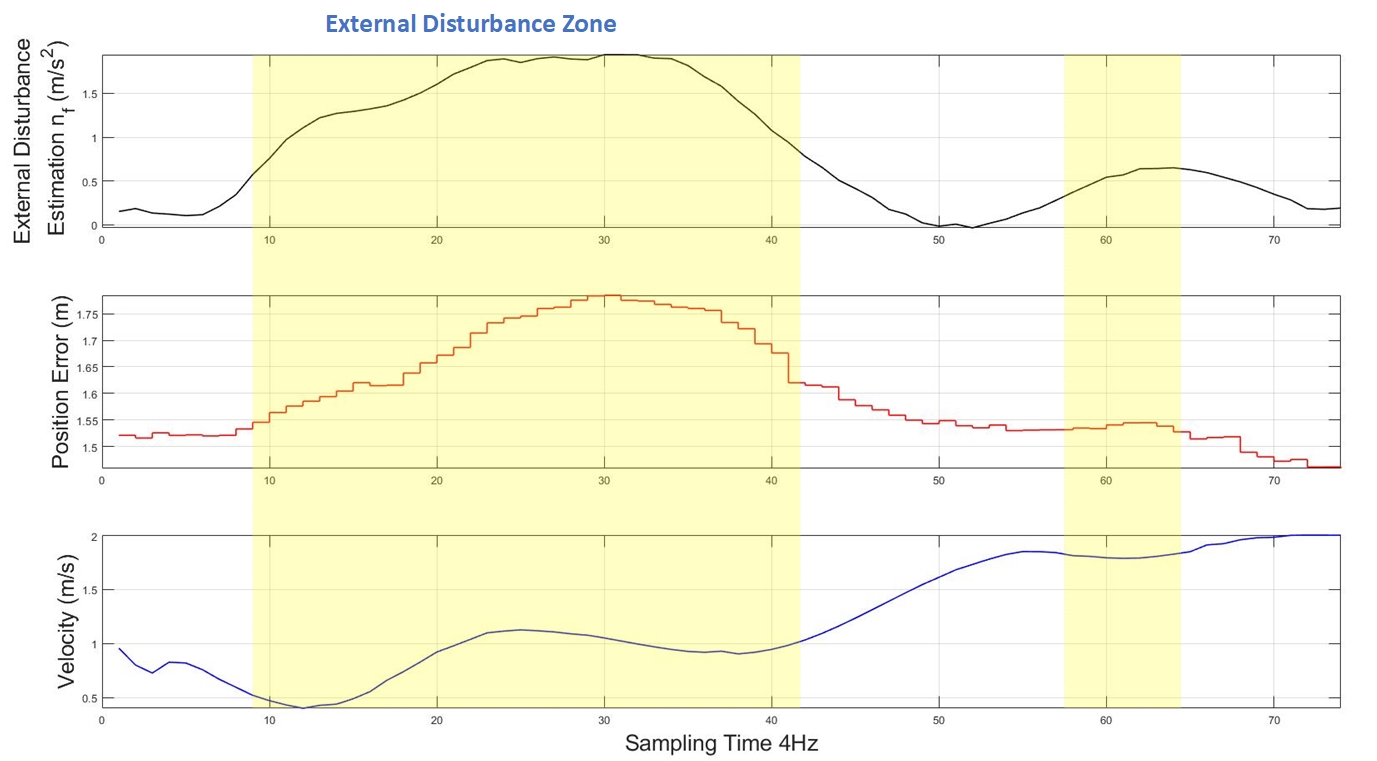}
  \caption{The simulation results: aerodynamic disturbance estimation $n_f (m/s^2)$, position error $(m)$ and velocity $(m/s)$.}
  \label{simulation_test}
\end{figure}

2) Comparative performance of KinoJGM trajectory optimization and tracking framework:
This simulation is based on the dynamic model described in \autoref{quadrotor_dynamics} and uses a Runge-Kutta integration to solve \autoref{mpcformulation} with $T_s = 50ms$ and time steps $N = 20$ \cite{wu2021external}. The implementation of GP-MPC mainly has two phases: training and prediction. In training, the data set for the GP model is collected by following aggressive trajectories \cite{torrente2021data} with the quadrotor. After the training set collection, we use the learning results to fit the GP-MPC. The training data set is collected with the velocity range $[-3,3]m/s$.

We compare our method against a state-of-the-art trajectory optimization and tracking algorithm, `Wu. \cite{wu2021external}', with dynamic aerodynamic disturbances added to our simulated environment. Wu.~\cite{wu2021external} uses a Kino-RS proposed by `Zhou. \cite{zhou2019robust}' to achieve route points and a Nolinear MPC (NMPC) to generate optimized trajectory and precise tracking. In \autoref{Comparison_of_tracking}, the second method, `Zhou. + GP-MPC', is a combination with Kino-RS algorithm (proposed in Zhou.~\cite{zhou2019robust}) and  GP-MPC. The results show that the GP-MPC tracking has better accuracy than NMPC in tracking. Without consideration for the aerodynamic disturbance, the second method has the lowest success rate and largest latency of the test scenarios. We can also see our KinoJGM has a greater success rate, efficiency and accuracy when handling large and complex aerodynamic disturbances. \autoref{simulation_env} shows that our proposed KinoJGM framework has smaller position errors (yellow areas) than Wu et al. \cite{wu2021external}. The aerodynamic disturbance estimation $n_f$, position error and quadrotor velocity in our simulated environment are shown in \autoref{simulation_test}.

\begin{table}[]
\caption{Comparison of Trajectory Optimization and Tracking}
\label{Comparison_of_tracking}
\begin{center}
\setlength{\tabcolsep}{1.7mm}{
\begin{tabular}{|l|l|c|c|c|}
\hline
Ex. disturbance                        & Method      & Succ. Rate & Time (s) & Err. (m) \\ \hline
\multirow{3}{*}{{[}0.0, 2.0, 0.0{]}}  & Wu. \cite{wu2021external}         & 90\%              & 9.98     & 5.17                        \\ \cline{2-5} 
                                      & Zhou. + GP-MPC & 80\%               & 13.61     & 4.1                         \\ \cline{2-5} 
                                      & KinoJGM    & \bf{90\%}             & \bf{7.32}     & \bf{1.60}                        \\ \hline
\multirow{3}{*}{{[}-1.0, 2.0, 0.0{]}} & Wu.~\cite{wu2021external}        & 75\%               & 14.3     & 6.11                        \\ \cline{2-5} 
                                      & Zhou. + GP-MPC & 70\%              & 13.8     & 5.33                        \\ \cline{2-5} 
                                      & KinoJGM    & \bf{80\%}              & \bf{9.76}     & \bf{2.19}                        \\ \hline
\end{tabular}
}
\end{center}
\end{table}


\section{CONCLUSIONS}

In this paper, we propose a complete online trajectory planning and tracking framework for unknown environments with dynamic disturbances. The framework integrates route search, trajectory optimization and tracking algorithms to overcome the effects of aerodynamic disturbances. We adopt a kinodynamic route searching method to improve our framework's efficiency whilst maintaining optimality. Using GP to augment the noise on our dynamic model of the quadrotor, an MPC significantly improves the accuracy of position tracking. We incorporate aerodynamic disturbance into the entire framework, both in planning and tracking. We demonstrate that our proposed approach can generate a trajectory efficiently and track it precisely, and that our approach compares favourably with selected benchmarks, i.e.,~\cite{zhou2019robust} and~\cite{wu2021external}, in simulation. 
Our ongoing work concerns demonstrating KinoJGM in real-world tests. We will also investigate adaptive methods for aerodynamic disturbance control, such as Bayesian Optimization MPC and Reinforcement Learning.






\bibliographystyle{unsrt}
\bibliography{ref}

\end{document}